\documentclass[letterpaper]{article} % DO NOT CHANGE THIS
\usepackage{aaai23}  % DO NOT CHANGE THIS
\usepackage{times}  % DO NOT CHANGE THIS
\usepackage{helvet}  % DO NOT CHANGE THIS
\usepackage{courier}  % DO NOT CHANGE THIS
\usepackage[hyphens]{url}  % DO NOT CHANGE THIS
\usepackage{graphicx} % DO NOT CHANGE THIS
\usepackage{natbib}  % DO NOT CHANGE THIS AND DO NOT ADD ANY OPTIONS TO IT
\usepackage{caption} % DO NOT CHANGE THIS AND DO NOT ADD ANY OPTIONS TO IT
\usepackage{algorithm}
\usepackage{algorithmic}
\usepackage{amssymb}
\usepackage{amsmath}
\usepackage{subfig}
\usepackage{caption}
\usepackage{subcaption}
\usepackage{graphics}
\usepackage[T1]{fontenc}

\usepackage{newfloat}
\usepackage{listings}
\DeclareCaptionStyle{ruled}{labelfont=normalfont,labelsep=colon,strut=off} % DO NOT CHANGE THIS
\floatstyle{ruled}
\newfloat{listing}{tb}{lst}{}
\floatname{listing}{Listing}
\title{Tree-Based Reconstructive Partitioning: A Novel Low-Data Level Generation Approach}
\author {
    Emily Halina,
    Matthew Guzdial
}
\affiliations {
    Department of Computing Science, Alberta Machine Intelligence Institute (Amii)\\
    University of Alberta, Edmonton, Alberta, Canada\\
    ehalina@ualberta.ca, guzdial@ualberta.ca
}

\begin{document}

\maketitle

\begin{abstract}
Procedural Content Generation (PCG) is the algorithmic generation of content, often applied to games. 
PCG and PCG via Machine Learning (PCGML) have appeared in published games. 
However, it can prove difficult to apply these approaches in the early stages of an in-development game. 
PCG requires expertise in representing designer notions of quality in rules or functions, and PCGML typically requires significant training data, which may not be available early in development.
In this paper, we introduce Tree-based Reconstructive Partitioning (TRP), a novel PCGML approach aimed to address this problem.
Our results, across two domains, demonstrate that TRP produces levels that are more playable and coherent, and that the approach is more generalizable with less training data. 
We consider TRP to be a promising new approach that can afford the introduction of PCGML into the early stages of game development without requiring human expertise or significant training data.

\end{abstract}

\section{Introduction}
% para 1: introduce pcg, introduce low example problem, motivate why this is a problem we want to solve
% introduce pcg and what it's been used for in the past
Procedural Content Generation (PCG) refers to the algorithmic generation of video game content \cite{shaker2016procedural}. 
By utilizing human-authored examples and heuristics, PCG approaches can generate many types of game content, from 3D models of trees \cite{speedtree} to entire game worlds \cite{minecraft}.
% we especially care about level generation
In particular, the task of level generation has been well explored across multiple games, including platformers \cite{awiszus2020toad}, dungeon crawlers \cite{van2013procedural}, and puzzle games \cite{guzdial2021impact}.
% and we ESPECIALLY care about level generation for active game development!
Many games have drawn on PCG to generate level structure live within a game or offline during development.
% Many games use PCG to generate content in an online or live setting, and there has been prior work in both industry and research on development tools that utilize PCG \cite{}.
However, many games do not employ PCG, or have found it negatively impacts development~\cite{Schreier_2017}.
% This includes the development of co-creative tools that allow designers to directly collaborate with Artificial Intelligence (AI) agents to create levels for existing games \cite{guzdial2019friend, halina2022threshold}.

% okay but here's da problem
A major challenge developers face in integrating PCG approaches into game development is generating quality content with minimal human authoring.
%Here's where the human authoring takes place examples like WFC or other PCGML approaches, components and rules for a constructive system, or functions for a search-based system.
The human-authoring required for PCG approaches ranges from example levels for Wave Function Collapse (WFC), to components and rules for constructive systems, or functions for Search-Based PCG (SBPCG).
The pace of game development is blistering, with rapid ideation and prototyping often leading to major changes in game balance and mechanics throughout the development process, potentially requiring frequent re-authoring of these human authored components \cite{ramadan2013game}.
This makes more human-authoring-intensive PCG approaches a burden, potentially sinking some game development projects \cite{Schreier_2017}.
An ideal system would generate quality, novel content with minimal human-authoring required.
%As such, for a PCG system to remain useful in such an environment, it must be able to produce quality output with limited human-authored examples or input for guidance, as at any point in development there may not exist more than a handful of human authored examples to reference.

% para 2: talk about other approaches, how they dont yet solve this problem
% Despite this, the majority of existing PCG methods struggle to produce quality output without adequate training data.
While there have been a number of PCG methods which can generate levels using minimal human-authored examples, these methods may not be generally appropriate for active game development.
% WFC: can work well but struggles with generation requiring more global constraints. can do the hand authored thing (i.e. sturgeon) but designers don't really think like that, also what if there are a lot or they constantly change
WFC can generate content from only a single human-authored example, but struggles with generating levels that adhere to global constraints \cite{karth2017wavefunctioncollapse}.
This makes it difficult to ensure WFC creates playable or soundly constructed levels, especially for game domains which have multiple requirements for playability, such as requiring both a key and door in a level.
A potential workaround to this problem is to hand-author these constraints, as in Sturgeon \cite{cooper2022sturgeon}.
%However, the hand-authoring of constraints can be impractical if they are numerous or constantly change, and designers may find it difficult to express their design goals in the language of constraints.
However, frequent re-authoring of these constraints may be burdensome, and designers may struggle to express their design goals in the language of constraints.
Similarly, SBPCG methods utilize functions which are faced with the same issue of requiring re-authoring, as evidenced by the lack of mainstream games using SBPCG in their development.

% Similarly, while search-based PCG approaches technically require minimal human-authoring, these approaches may be difficult for designers, as they must represent design knowledge as functions.
% Search-based PCG approaches technically require minimal human-authoring in the form of guiding heuristic functions. 
% However, these approaches may be similarly difficult for designers, as they must represent qualitative design knowledge into functions that describe output quantitatively.

% Cite the Summerville and Snodgrass comparison of different amounts of training data for PCGML approaches
%Other PCG approaches can be run with minimal training data such as Markov Chains and Long Short-Term Memory deep neural networks (LSTM).
%However, these methods have previously been shown to heavily plagiarize when run with few examples, and the expressive volume of the two models is greatly reduced \cite{snodgrass2017studying}. 

% para 3: but our approach does! outline the steps of the approach
% high level intuitive argument about how this can help solve this problem
In this paper, we introduce a new PCG method based on Monte Carlo Tree Search (MCTS) and Space Partitioning that can generate a suite of levels from as little as a single human-authored example.
On a high level, our approach works by first extracting information from the search tree of one or more MCTS playthroughs of a source level.
Using this information, we then reconstruct an intermediary binary representation of the level approximating the source geometry.
Finally, we utilize both the source level and the MCTS search tree to perform a binary space partition and probabilistic threat placement to complete the level.
We dub this new method \textbf{Tree-based Reconstructive Partitioning} (\textbf{TRP}) after these three high-level steps.
Besides the single human-authored level example, TRP requires access to a forward model of the game and a small amount of domain-specific knowledge provided by a human designer.
Notably, TRP's domain-specific knowledge is not a value or heuristic function, in which a human designer must determine how to evaluate the goodness of a level, but instead a set of affordances and ranges which constrain generation. 
As such, TRP could feasibly be used in scenarios including early development when training data is scarce.

% para 4: contributions! outline? (maybe skip maybe add we have lots of space)
The contributions of this work to the game AI community  are as follows:
\begin{itemize}
\item Tree-based reconstructive partitioning (TRP), a novel PCG method that generates valid, unique levels from a single example.
\item An empirical evaluation of TRP's performance in comparison to six baseline PCG approaches across two disparate game domains. 
\item A brief discussion of the potential extensions and applications of TRP to a variety of problem domains.
\end{itemize}

\section{Related Work}
In this section, we overview prior work on PCG in the context of level generation, and provide a background primer on MCTS and its prior use in other PCG approaches.

\subsection{PCG for Level Generation}
PCG approaches can be grouped into categories with similar methodologies.
One such category is classical PCG, approaches that do not use machine learning, including constructive, search-based, and constraint-based PCG \cite{charity2020mech, horswill2021answer}.
These methods require a human-authored notion of goodness to guide generation, such as rules or components for constructive PCG, functions for SBPCG, and constraints for constraint-based PCG.
While some of these approaches are commonly used within completed games running PCG live, these requirements make them difficult to develop from scratch during early game development \cite{Schreier_2017}.
Procedural Content Generation via Machine Learning (PCGML) refers to the family of PCG approaches that utilize machine learning techniques \cite{summerville2018procedural}.
While the advent of deep learning has brought about many powerful PCGML approaches, we focus specifically on approaches that require a small amount of data due to our specific use-case.
One such approach is Wave Function Collapse (WFC), which can use patterns from a single example level to constrain generation \cite{karth2017wavefunctioncollapse}.
WFC, as previously stated, can struggle to adhere to global constraints such as playability without potentially intensive human-authored assistance \cite{cooper2022sturgeon}.
Other PCGML approaches can run with minimal training data such as Markov Chains, Long Short-Term Memory deep neural networks (LSTM), and Generative Adversarial Networks (GAN) \cite{snodgrass2013generating, summerville2018procedural, awiszus2020toad}.
However, prior work shows that these methods can heavily plagiarize when run with few examples, with the expressive volume of the models limited by the sample size \cite{snodgrass2017studying}. 

% pcgrl
% tehcnically requires no training data, but same problem as search based (needs reward function)
Procedural Content Generation via Reinforcement Learning (PCGRL) is an emerging group of approaches that utilize reinforcement learning to generate game content \cite{khalifa2020pcgrl}.
These approaches work by formulating level generation as a Markov Decision Process (MDP), making incremental changes to a piece of content, and have been successfully applied to multiple domains \cite{earle2021}.
However, similarly to search-based methods, PCGRL approaches require a reward function, with the same limitations as SBPCG heuristics.
In addition, small changes to the game may require a complete reformulation of the MDP, which requires a large amount of expertise in reinforcement learning that game designers may lack \cite{berner2019dota}.
% Would be good to add another sentence here that while MCTS might be an example of model-based RL we don't consider this PCGRL since the MCTS portion isn't doing the generation and we could swap out MCTS for another approach to generate a tree representing paths through the level.
While MCTS is an example of model-based RL, we don't categorize our approach under PCGRL, as MCTS doesn't directly alter level structure.

\subsection{Paths and Trees in PCG}

Prior PCG approaches have made use of paths and trees for data augmentation and as a part of the level generation process.
Exhaustive PCG exhaustively explores a search tree of possible designs for a level \cite{guzdial2021impact}.
This is distinct from our approach, which uses a search tree from a playthrough of the level as input rather than as a part of the generation process.
Existing PCG approaches use paths for a variety of tasks, including learning level topology and performing level blending across game domains \cite{summerville2015learning, sarkar2020exploring}.
Our approach differs in that we utilize an entire search-tree, including branches which ended in failure states.

\subsection{Monte Carlo Tree Search}
% talk about MCTS! and the approaches that use it (mcmcts, kynan, angry bords)
% mention that you could use other tree search algorithms / connect to epcg
Monte Carlo Tree Search (MCTS) is a stochastic tree search algorithm that simulates future actions to find the best next choice \cite{browne2012survey}.
MCTS has proven to be a strong approach for automated game playing, both alone and embedded within larger systems \cite{chaslot2008monte, fu2016alphago}.
In the context of game playing, MCTS works by building a tree of possible future game states, with nodes represents game states and edges representing actions taken by the player.
%<Sentence here about how we use this tree as the input for our generative approach.>
Our approach uses this tree as input for the generative process.
Over each iteration, MCTS incrementally evaluates the value of each node representing a future state, which in turn can be used to infer a good next action for the current state.
This process can go on for as many or as few iterations as desired, allowing for a balance between performance and computational requirements.
Each iteration of MCTS is made up of the following four steps.
\subsubsection{Selection} 
From the root node, we successively select next nodes. We make this selection based on a tree policy function, which guides where next to explore. We stop when we reach a node with an unexpanded child. 
This tree policy guides how MCTS iteratively builds the tree from the root, balancing exploration of new states versus exploitation of known good states.
\subsubsection{Expansion}
We randomly select an unexpanded child of the selected node from the Selection step.
\subsubsection{Simulation}
The purpose of the simulation step is to simulate the long-term result of the game from the state of the unexpanded child.
In our case, we simulate a playthrough beginning from this unexpanded child, taking random actions in every step. 
This playthrough continues until either a pre-defined number of actions are taken, or we reach a terminal state, such as a win, loss, or draw.
\subsubsection{Backpropagation}
We evaluate the value of the final state we reach in the random playout, then backpropagate this value back up the tree to each parent until we reach the root.

\subsubsection{MCTS in PCG}
While MCTS has been incorporated into some prior PCG systems, it has mostly been used to directly guide the generation process, such as in Summerville et al.'s MCMCTS where MCTS places columns of existing levels to create new levels \cite{summerville2015mcmcts, graves2016procedural}. MCTS has also been used as part of a heuristic in SBPCG approaches \cite{sorochan2022generating}.
This is in contrast to our approach, which uses MCTS to play through a source level to derive a tree that serves as input to a generation process.

\section{Tree-based Reconstructive Partitioning}

\begin{figure}
    \centering
    \includegraphics[width=\columnwidth]{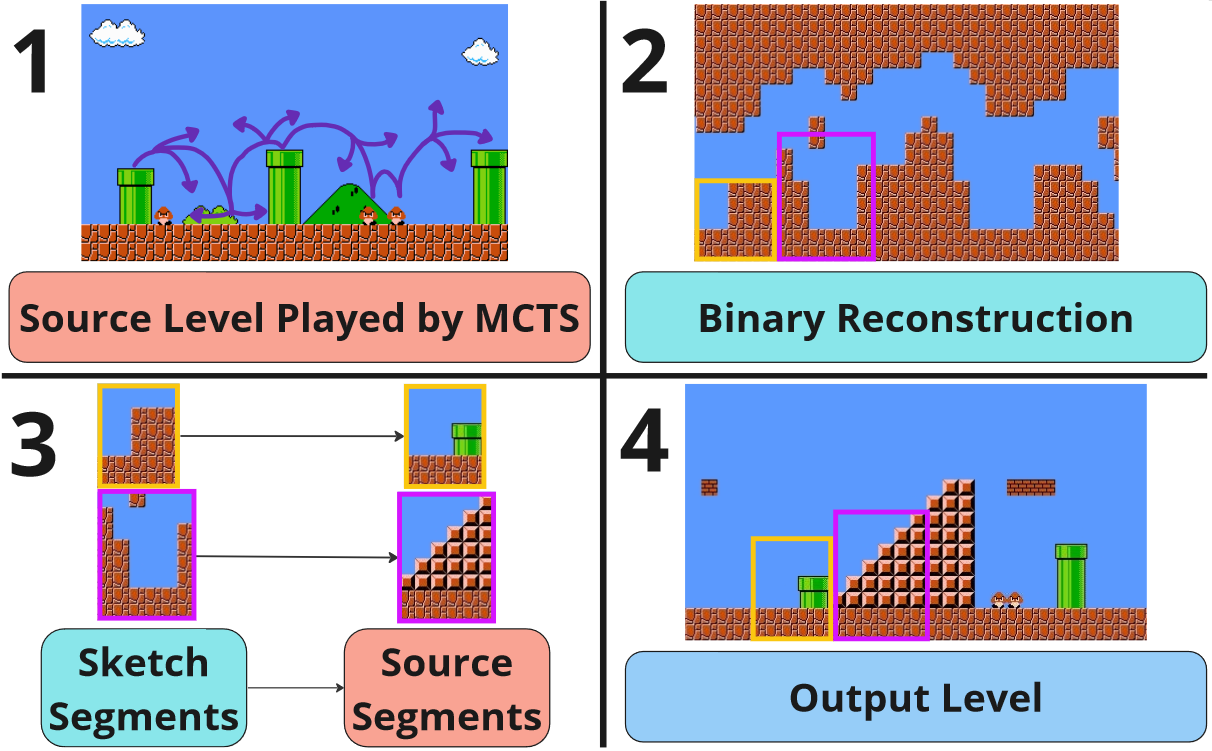}
    \caption{Visualization of the TRP generation pipeline. With MCTS search trees and designer knowledge kit as input, TRP constructs an intermediate binary representation. TRP then uses the input data to perform a binary space partition and probablistic threat placement, producing the final level.}
    \label{fig:sysoverview}
\end{figure}

In this section we discuss the implementation of our approach, Tree-based Reconstructive Partitioning (TRP), in terms of a generic, token-based game domain.
Figure \ref{fig:sysoverview} depicts our level generation pipeline from the original source level to our generated level as output.
As input, TRP takes in the source level and a ``knowledge kit'' of domain-specific human-authored information. 
This information, described in full below, takes the form of affordances, ranges, and sets of tokens, such as a group of tokens that should be considered ``threats.''
First, we perform one or more playthroughs of the source level with a tree search algorithm, saving the search trees from these playthroughs.
In this paper, we use MCTS.
We then convert the output search trees to an intermediary binary representation.
Finally, we perform a binary space partition using this intermediary representation to generate new level structure, and use the position of deaths in the search tree to populate our new level with threats.

\subsection{Application Requirements}
% we need a token based rep!
TRP assumes a discrete token-based representation of the game domain's levels, as a discrete representation is required to perform binary space partition.
% if you dont have one you can probably get one!
%Notably, a game without a ``native'' token-based representation, such as a 3D game, could be discretized to use this approach, as in prior work encoding 3D environments as graph structures for generation \cite{giacomello2018doom}.
% we need a forward model!
Further, TRP requires access to a forward model of the game and a game-playing heuristic in order to collect the search tree information.
% if you dont have one wait for our future work!
While this is a somewhat burdensome requirement, we speculate that the use of an engine or plugin that automatically sets up a forward model for the developers could alleviate the authoring work required to add this functionality.

\subsection{Tree-based Reconstruction}
% we play through using MCTS! here's why MCTS
As input, TRP requires one or more search trees representing potential paths through a game level.
Thus, we begin by conducting one or more playthroughs through the source level using MCTS to guide the player.
% it's efficient for devs
We chose MCTS due to its ``just-in-time'' nature which allows the algorithm to perform well on a variety of hardware strengths. 
This is a relevant concern for developers who may not have access to a large amount of computing power.
% also it's repeatable: maybe this one should come first?
Further, the stochastic nature of MCTS allows us to explore different possibilities throughout different playthroughs, i.e., the player character taking a different choice in a branching path.
% btw, this doens't haaave to be MCTS
However, any tree search algorithm could be used for this stage, as long as it outputs the required search tree information.

% here's how MCTS is guided
For both of the domains in this paper, we used the UCT selection policy, where the value of the $i$-th node is given by
\begin{equation}
    v_i + c \sqrt{\frac{\ln N_i}{n_i}}
\end{equation}
%where $v_i$ is the average reward value of the $i$-th node, $c$ is a tunable constant, and $N_i$ and $n_i$ are the number of visits to the $i$-th node's parent and the $i$-th node respectively.
where $v_i$ is the average reward value, $c$ is a tunable constant, and $N_i$ and $n_i$ are the number of visits to the node's parent and itself respectively.
% i talk about the rollout depth and value of c in the knowledge kit section because it felt most natural
The reward value $v_i$ of a given state is domain-specific, as it is dependent on the designer's description of goal states, as discussed in subsection 3.4.

% here's the info we log through these playthroughs
Throughout the playthroughs, we retain two key pieces of information: the position of the player at each timestep, and the position and cause of any failure states (i.e. death) within any of the rollouts.
We utilize the player positions across all the playthroughs to construct an intermediary binary representation of tokens that represent whether there is structure at a position or not.
Figure \ref{fig:sysoverview} contains examples of this intermediary representation of the combined paths of each search tree through a source level.
The number of playthroughs performed through the level is another parameter users can change to affect the generated artifacts, as it impacts this intermediary representation.
The intention behind using the player's positions to generate a level is to roughly approximate the geometry of the target level while finding the \textit{most important} required elements to progress.
If we can generate a level with a similar path, we hypothesize that the design intention that afforded that path in the original level may be captured in the new level.
This intermediary representation is used in the following steps of generation alongside the source level in order to create the final output.

\subsection{Binary Space Partitioning}
% motivating why we want to do partitioning in the first place (make it look like a real level while also retaining the information we extracted with the tree reconstruction) 
We fill in the intermediary binary reconstruction of the source level in an attempt to aesthetically match the content of the source level while retaining the extracted structure from the trees' paths.
To do this, we use an example-driven binary space partition adapted from Snodgrass' implementation \cite{snodgrass2019levels}.
Binary space partition is an algorithm for constructing levels by filling in a binary ``sketch'' of a level with corresponding pieces of a source level.
For our use-case, we use the intermediary reconstruction as the binary sketch, and the original level stripped of all threats as the source level.
% i give this domain specific example here to relate to why we wanna use this!
We chose binary space partition for this task due to its ability to retain structural information from the reconstruction while replicating patterns from the original source level, such as complete pipes in Super Mario Bros.
% rather than providing an algorithm in blocks, i'm just going to do this: if you think it'd be better for me to make an algorithm thingy, i can do that, but it'd basically the same as the one in snodgrass' paper so i figured i'd just cite it and describe it in plaintext since it takes like 2 sentences

Binary space partitioning works by first dividing the binary sketch into segments of size $w \times h$. 
$w$ and $h$ vary in size, but are bounded by a parameter $s$ that limits their maximum value.
Next, each of these segments are ``matched up'' with corresponding segments from the source level by a binary similarity function which counts the number of similarities between the two. This function is given by
\begin{equation}
    \sum_{(i, j) \in |S_{n \times m}|} 1 - \text{sign}^2(S[i, j] - L[i, j]) 
\end{equation}
where $S$ and $L$ are $n \times m$ segments of level, with $S[i, j] = 0$ if the token at the $(i, j)$-th position of $S$ is an empty token and $S[i, j] = 1$ otherwise.
Sign is the sign function, which outputs $1$ and $-1$ for positive and negative numbers respectively, and $0$ otherwise.
We use a sliding window across the source level to find all of the closest matching segments of the same size, then randomly select one of the resulting matches with uniform probability.
These matches ``fill in'' the details of the sketch one at a time until all of the matches are processed.
As an example using $s=5$, if a binary sketch was divided into 4x3 and 2x5 segments, the algorithm would first examine all 4x3 segments of the source level, matching the closest one via Equation 2.
Then, the algorithm would repeat this process for the 2x5 segments.

% enemy time
After completing the partitioning, our final step is to repopulate our output level with threats.
% the probabilistic thing we did, and why we did it this way
Our threat repopulation methodology leverages the information from the collected search trees to identify the most relevant threats in the level, then places those threats in order from most to least relevant to reach a designer-set desired difficulty threshold.
%Relevance is determined by the percentage of deaths caused by a threat in a given position with respect to all of the deaths across all of the rollouts in the search trees.
% Relevance is determined through the formula
% \begin{equation}
%     \frac{d_e}{d_t}
% \end{equation}
% where $d_e$ represents the number of deaths caused by a threat $e$ in a given position with respect to all rollouts, and $d_t$ is the number of total deaths across all rollouts.
Relevance is determined through the formula $\frac{d_e}{d_t}$, where $d_e$ represents the number of deaths caused by a threat $e$ in a given position with respect to all rollouts, and $d_t$ is the number of total deaths across all nodes.

The identity of a threat is domain-specific and the difficulty threshold is an adjustable parameter, as explained in subsection 3.4 below.
Threats are placed directly within the corresponding position of the failure state in the search tree.
We chose this methodology in order to maximize the impact and intentionality of the placement of threats by the system in an attempt to replicate the patterns of threat placements within the original level.
By placing only the most relevant threats for a given level, we hope to mimic the designer's original intentions with threat placement and create levels which provide the desired level of challenge without overpopulating threats throughout the level.

\subsection{Knowledge Kit}
TRP's required human-authoring takes the form of a set of human-authored affordances and ranges.
This information is used to guide the tree search to play through the game, and help with the categorization and repopulation of threats.
These parameters include three affordances regarding the game's states and entities, two numerical parameters for MCTS, and three numerical parameters for TRP which affect generation.

% we need goals/failures, and enemies!
The three human-authored affordances are the definition of a sequence of goal states $G = \{S_{g_1}, \dots, S_{g_k}\}$, a sequence of failure states $F = \{S_{f_1}, \dots, S_{f_k}\}$ and a list of tokens which represent threats within the game domain.
The first of these guides the tree search agent through the level playthroughs, as the value of states relies on their proximity to the current goal.
Designers can provide an arbitrary number of ordered goal definitions, such as first collecting a key and then reaching a locked door, which would be encoded as a sequence of two goal states $\{S_{g_k}, S_{g_d}\}$.
% hey but what does that actually look like in terms of value
In both domains, we computed reward value based on the Manhattan distance from the closest current goal state to improve the MCTS agent's performance.
% if this doesn't make snese look at the concrete example that will exist below!
%An example of this weighting is described in the Knowledge Kit of the GVGAI Zelda domain below.
Similarly, the failure states corresponding to the current goal state determines negative reward signal for the MCTS agent.
The list of threat tokens is used to identify the causes of failure states during the threat repopulation phase based on token proximity.
%(make an argument here about why this is different then requiring a SBPCG heuristic or PCGRL reward function)
Notably, these requirements are fundamentally different from the requirements of SBPCG systems, as we require designers to provide information about game states and entities instead of level quality.

% Notably, these requirements are fundamentally different from the requirements of search-based PCG systems, which require hand-authoring of heuristic functions that determine the quality of a given level or state.
% Instead, these requirements only require designers to provide information about the game's states and entities, which we hypothesize will make the approach more approachable for designers without explicit PCG experience.

% MCTS parameters
There are two numerical parameters for MCTS that designers can adjust from their default values to improve both computational performance and the strength of the agent as a player.
These are the tuneable exploration constant $c$ defined in subsection 3.2, and the maximum depth for rollouts in the Simulation step.
The theoretically optimal value for $c$ is $\sqrt{2}$ \cite{kocsis2006bandit}.
However, adjusting it may improve performance in some domains, allowing the agent to play more complex or challenging levels.

By default, the maximum depth for rollouts is unbounded (i.e. a rollout will continue until a terminal state is reached), but this may not be appropriate for more complicated domains where it may take hundreds or thousands of steps to reach a terminal state.
Instead, by bounding the depth of rollouts, we can force the MCTS agent to optimize more locally, which can improve performance in some domains \cite{zook2019monte}.
% ok but why is it feasible that we ask designers to do this! (maybe there could be a stronger argument here?)
%While these parameters do require tuning in order to maximize the potential of the MCTS agent, this tuning does not require designers to dive into the implementation details of the algorithm, and simply can be performed in a ``guess and check'' fashion.
While these parameters do require tuning in order to maximize the potential of the MCTS agent, we found the set of reasonable parameters to be quite similar across both our evaluation domains.
Further, in a potential designer-focused TRP tool, these parameters could be automatically adjusted and/or discovered via a parameter optimization approach such as a grid search.

% Our own parameters!
TRP takes three numerical parameters which allow human designers to control the generation process.
The first, $t \in \mathbb{N}$, denotes the number of playthroughs TRP will perform of the source level.
This count affects the openness of the intermediary level, with more playthroughs opening up more options and identifying more paths in the source level.
The second, $s \in \mathbb{N}$, denotes the maximum allowed size of any segment taken from the source level during the binary space partition, affecting the amount of direct plagiarism tolerated in the output level.
The third, $e \in [0.0, 1.0]$ denotes the weighted percentage of threats to take from the search tree as explained above.
This affects the resulting threat density of the output level, with $0.0$ representing no threats, and $1.0$ representing the maximum density of threats.

While TRP is usable with fixed parameters, having these parameters vary over a range dramatically increases the diversity of the generated levels.
As such, for each domain, we present results with both a fixed set of parameters and a predefined range, with the knowledge kits for both of our evaluation domains provided in the following section.
% so we totally just picked these values ourselves, but it was easy to try things and figure it out (and it would be easy for a designer to punch in some numbers especially with an intuition of what they do and go from there)
For both of our domains, we determined the parameter values by our own human judgement of the generated levels.
Due to the speed of generation, it was trivial to determine reasonable values for both domains, taking minutes of experimentation.
%This was a relatively quick and simple procedure in both domains due to the speed of generation, taking a handful of minutes of experimentation with different values.

%%%%%% 
%applied all of the edits so far, along with some of kristen's suggestions that i agreed with!
%%%%%%

\section{Domains}
% why we pick these ones 
% what the implementation looked like (engine, knowledge, etc)
In this section, we discuss our implementation of TRP within two game domains: Super Mario Bros. and GVGAI Zelda \cite{Khalifa, perez2016general}.
We chose Super Mario Bros. as a domain as it is a ubiquitous standard for PCG approaches \cite{summerville2018procedural}.
We chose GVGAI Zelda as it directly contrasted Super Mario Bros. in gameplay and generation requirements, and represented a domain in which levels are encoded in very few tokens.
In the following subsections, we discuss the implementation details of each domain and provide the knowledge kit of human-authored information used to generate levels for them.
We provide the source code for each of these implementations\footnote{https://github.com/emily-halina/TRP/}, including scripts for generating new levels from prior MCTS playthroughs.

\subsection{Super Mario Bros.}
% okay what's this game and what's the goal
Super Mario Bros. is a platformer game originally released for the NES in 1983 \cite{miyamoto1985super}.
The player starts on the far left of each level, and the goal is to navigate to a flagpole at the end of a stage using jumps, movement, and various powerups.
An example of a Super Mario Bros. level is depicted in Figure 2.
% what engine did we use? how's it set up? our two pre-reqs are covered (token rep and forward model)
Our implementation uses version 0.80 of the updated Mario AI Framework, a Java reimplementation of the original Super Mario Bros.~\cite{Khalifa}.
Each level is encoded as an $n \times 16$ matrix of tokens within a text file, with each token representing a singular tile or block in the game.
The framework includes a forward model, which we used to implement the MCTS agent required by our approach.

% what's the knowledge kit for this domain
\subsubsection{Super Mario Bros. Knowledge Kit}
% goals and failures
We defined the goal state in Super Mario Bros. to be when Mario's x-coordinate is the same as the flagpole's.
Notably, this goal state may correspond to several possible world states.
We defined the failure states as states in which Mario is in contact with an enemy or enemy projectile, or is one tile below the screen, which represents Mario falling in a pit.
% pointing out that this is easy for developers to do and natural
%These two sets are naturally defined, as they correspond directly with in-game success and failure.
Using these two sets and the previously defined scheme of measuring the Manhattan distance from the goal, the value function guiding MCTS scores nodes based on how far to the right Mario can reach.
This is similar to Jacobsen et al.'s prior work on implementing MCTS for Super Mario Bros. \cite{jacobsen2014monte}.   
% tokens that represent threats
The set of threat tokens were simply any token that is associated with an enemy, such as Goombas, as well as any air tile which is at the very bottom of the screen, which represents the bottom of a pit.
Notably, we were able to perform this encoding by swapping out the bottom tile for another special token in a pre-processing step.
This gave us a total of $17$ threat tokens in this domain.

% parameters for MCTS 
For MCTS parameters, we used an exploration constant value of $c = 0.25$, and a maximum rollout depth of $12$.
These values are based on both prior work on MCTS controllers for Super Mario Bros., and our own empirical evaluation \cite{jacobsen2014monte}.
With larger rollout depths, Mario becomes noticeably ``cowardly,'' avoiding all threats, and as such we keep the depth relatively low.

% parameters for TRP
The parameters for the fixed and variable versions of TRP are as follows.
The fixed version uses parameter values 
\begin{equation}
    (t = 2, s = 9, e = 0.67)
\end{equation}
and the variable version uses parameter ranges 
\begin{align}
    &t \in \{1, 2, 3, 4, 5\}\\
    &s \in \{8, 10, 12, 14, 16\}\\
    &e \in \{0, 0.33, 0.67, 1\}
\end{align}
with $t, s, e$ defined as above.
We chose $1$ through $5$ as a range for playthroughs ($t$), as we found including more than $5$ playthroughs began to show diminishing returns in terms of exploration of new paths.
We chose the values of $s$ based on the width of a $16 \times 16$ chunk of screen, with $8$ representing half a row or column, and $16$ representing a full row or column. We selected the values of $e$ to create levels of varying difficulty.

\subsection{GVGAI Zelda}
% okay what's this game and what's the goal
The General Video Game AI (GVGAI) framework is a research framework mainly used for games research competitions \cite{perez2016general}.
GVGAI is made up of a corpus of over $100$ single-player games, including GVGAI Zelda.
GVGAI Zelda is a top down puzzle game where the player navigates a tile-based environment full of enemies with orthogonal movement.
The game's goal is to collect a key then reach a door in the level.
Examples of GVGAI Zelda levels are depicted in Figure \ref{fig:zeldalevels}.
% what engine / implementation are we using? our two pre-reqs are covered (token rep and forward model
We utilize the OpenAI Gym implementation of GVGAI which acts as a Python wrapper for the Java implementation of GVGAI \cite{torrado2018deep}.
Each level is encoded as an $13 \times 9$ matrix of tokens within a text-file, with each token representing a singular tile or block in the game.
As the Gym implementation did not contain a forward model by default, we implemented our own which can be found in the provided source code.

\subsubsection{GVGAI Zelda Knowledge Kit}
We defined the sequence of goal states in GVGAI Zelda as $\{S_k, S_g\}$, which encodes the required steps of first obtaining a key, and then reaching the door in order to complete a level.
We defined the failure state as any state in which the player has the same x and y position as an enemy.
%Using the ordered set of goals, the MCTS agent is guided by trying to minimize the distance from the first unachieved goal, with a weighting applied based on the number of goals achieved to keep the value between $0$ and $1$.
% enemies
The set of tokens representing threats consists of the three enemy types within the game, which differ by moving randomly at different frequencies and intervals.
This gave us a total of $3$ threat tokens in this domain.

% mcts params
For MCTS paramaters, we used an exploration constant $c = 0.25$, and a maximum rollout depth of $3$.
These choices were informed by our own empirical evaluation of the performance of the agent.
While $3$ may seem like a relatively small depth for exploration, the randomized movement of enemies forces the agent to make decisions local to its immediate surroundings.

% trp params
The parameters used for the fixed and variable versions of TRP are as follows.
The fixed parameter values
\begin{equation}
    (t = 1, s = 2, e = 0.25)
\end{equation}
and the variable version uses parameter ranges
\begin{align}
    &t \in \{1, 2, 3\}\\
    &s \in \{1, 2, 3, 4, 5\}\\
    &e \in \{0, 0.25, 0.375, 0.5\}
\end{align}
with $t, s, e$ defined as above.
As before, we chose the range of playthroughs ($t$) based on observed diminishing returns from including more playthroughs.
For this domain, we found that smaller values for $s$ performed best, with larger values than $5$ resulting in complete plagiarism of the source level.
The set of values for $e$ is lower in comparison to Super Mario Bros., as the density of deaths is much higher in this domain.

\section{Evaluation}

\begin{figure}
  \centering
  \begin{tabular}{@{}c@{}}
    \includegraphics[width=\linewidth]{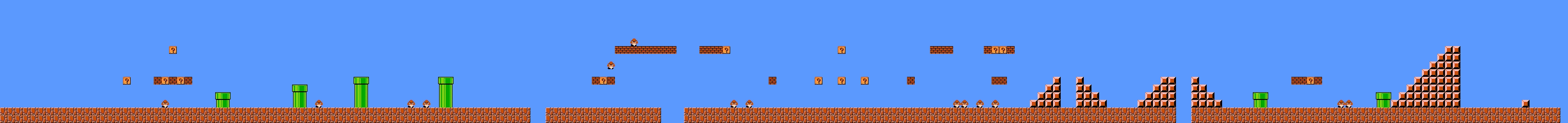}\\
    \small (a) Original 1-1
  \end{tabular}

  \begin{tabular}{@{}c@{}}
    \includegraphics[width=\linewidth]{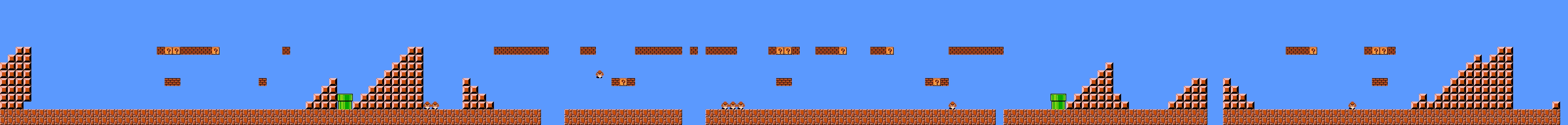} \\
    \small (b) \textbf{TRP}, $t=2, s=9, e=3$
  \end{tabular}
  
  \begin{tabular}{@{}c@{}}
    \includegraphics[width=\linewidth]{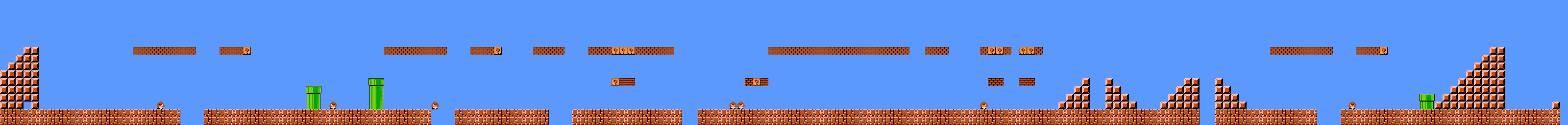} \\
    \small (c) \textbf{TRP}, $t=4, s=10, e=3$
  \end{tabular}

  \begin{tabular}{@{}c@{}}
    \includegraphics[width=\linewidth]{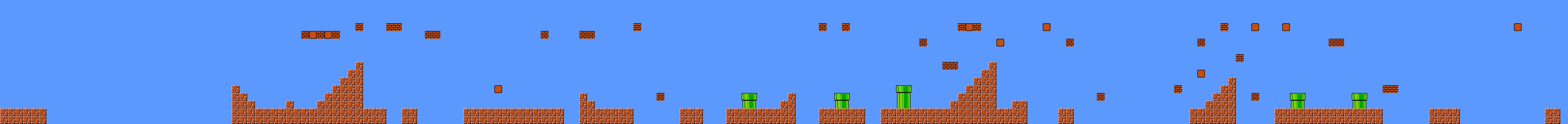} \\
    \small (d) Sturgeon (baseline)
  \end{tabular}

  \begin{tabular}{@{}c@{}}
    \includegraphics[width=\linewidth]{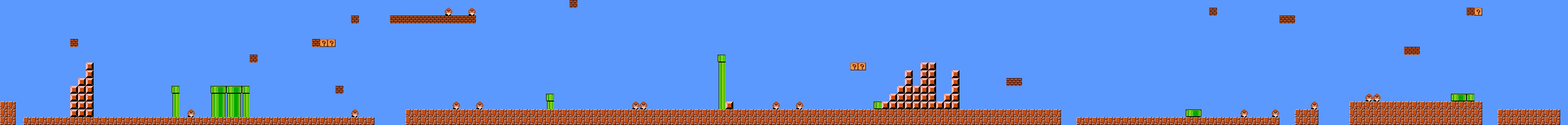} \\
    \small (e) Markov Chain (baseline)
  \end{tabular}

  \begin{tabular}{@{}c@{}}
    \includegraphics[width=\linewidth]{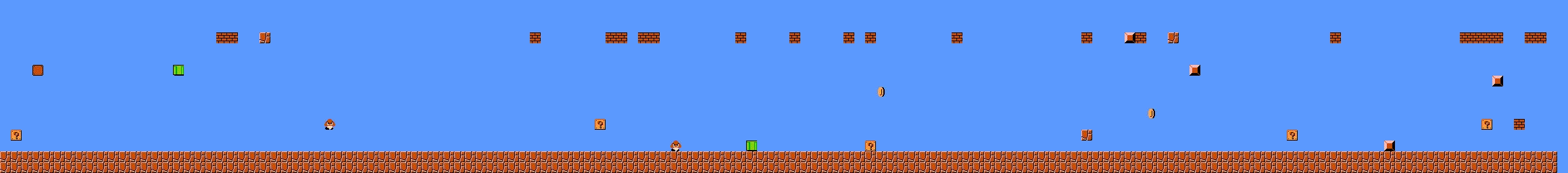} \\
    \small (f) Autoencoder (baseline)
  \end{tabular}

  \begin{tabular}{@{}c@{}}
    \includegraphics[width=\linewidth]{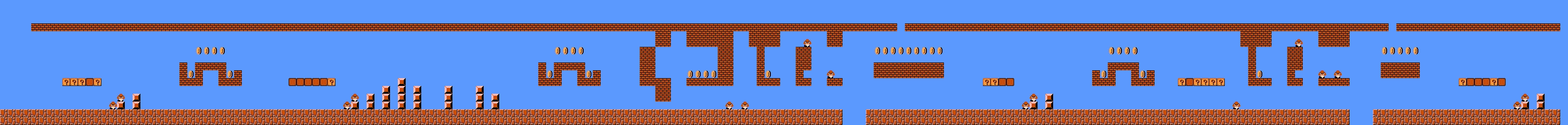} \\
    \small (g) MCMCTS (baseline)
  \end{tabular}

  \begin{tabular}{@{}c@{}}
    \includegraphics[width=\linewidth]{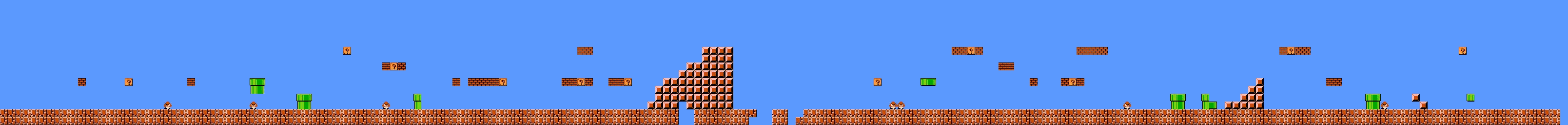} \\
    \small (h) TOAD-GAN (baseline)
  \end{tabular}

  \caption{Randomly selected levels generated by TRP and baselines for Super Mario Bros. using Level $1$-$1$ as training data, except Autoencoder and MCMCTS which use $1$-$1$ and $1$-$2$. For more examples, see the Github repository.}
  \label{fig:mariolevels}
\end{figure}

\begin{figure}
    \centering
    \begin{tabular}{cc}
        \includegraphics[width=0.4\linewidth]{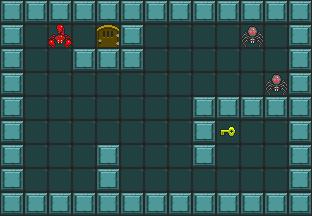} & 
        \includegraphics[width=0.4\linewidth]{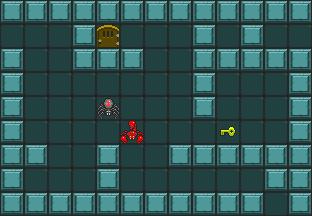} \\
        \small (a) Original Level 1 & \small (b)  \textbf{TRP}, $t=1, s=3, e=2$  \\
       \includegraphics[width=0.4\linewidth]{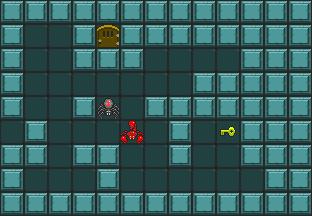}  & \includegraphics[width=0.4\linewidth]{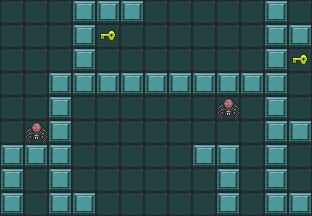} \\
       \small (c) \textbf{TRP}, $t=1, s=2, e=2$ & \small (d) WFC (baseline) \\
       \includegraphics[width=0.4\linewidth]{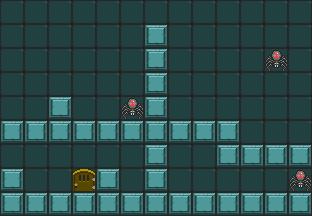}  & \includegraphics[width=0.4\linewidth]{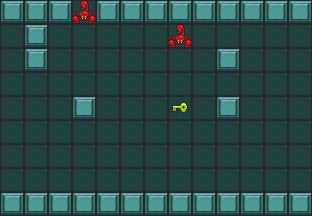} \\
        \small (e) Markov Chain (baseline)  &  \small (f) Autoencoder (baseline)
    \end{tabular}

    \caption{Randomly selected levels generated by TRP and baselines for GVGAI Zelda using Level $1$ as training data, except Autoencoder which uses Levels $1$ and $2$.}
    \label{fig:zeldalevels}
\end{figure}

% What are we trying to do? Why didn't we do X (study with game designers)
In this section, we discuss the evaluation of our approach in the previously defined game domains, overviewing the baselines and evaluation metrics.
Recall that our goal is to address the problem of producing high-quality content with minimal human-authoring.
We could have evaluated our approach through a human subject study with game designers either directly using TRP in practice or evaluating generated level content.
However, as this paper is an initial foray into TRP as an approach, our goal is to perform an exhaustive analysis of the approach that would not be appropriate for a human subject study.

% in a paragraph, what did we do (then expanded on below in the subsections)
% the baselines!
Instead, to evaluate our approach we compared levels generated by TRP to levels generated by a number of baseline approaches across both game domains.
These baselines were Markov Chains, Markov Chain Monte Carlo
Tree Search (MCMCTS), WFC, Sturgeon, an autoencoder-based PCGML approach, and TOAD-GAN.
We had two levels from each domain as human-authored training data, namely Level $1$-$1$ and $1$-$2$ from Super Mario Bros. and Level $1$ and $2$ from GVGAI Zelda.
% for the last time you cannot have search-based pcg i am turning this car around
Notably, we do not include a SBPCG or PCGRL baseline, as the required authoring of level quality functions makes it difficult to compare these approaches to those reliant on only a handful of human-authored examples.
The implementation details for each baseline are outlined in the Baselines subsection.

For comparative metrics, we employed playability, plagiarism of the original source levels, and self-similarity between the population of generated levels.
These metrics were chosen as stand-ins for a human evaluation of quality due to their approximation of the validity, originality, and variety of generated levels respectively.
The details of each metric are outlined in the Metrics subsection.
% overview each baseline approach
\subsection{Baselines}

\subsubsection{Markov Chain (MC)} Markov chains are a classic PCGML approach that involves using local context in order to probablistically determine which tile to place in a given location.
The probability distribution of tiles based on their immediate surroundings is learned from training examples, which in our case are the human-authored levels.
We chose to use Markov chains as a baseline due to their prominence in early PCGML level generation work, and their success in domains such as Super Mario Bros. \cite{summerville2018procedural}.
For both of our domains, our implementation uses a $2 \times 2$ context window where the value of the top-right tile is based off the remaining three tiles surrounding it in an ``L'' shape with equal weighting.
This window size was chosen to maximize the amount of training examples given the small sample size, as larger window sizes resulted in many ``unseen states'' during generation.
In our implementation, we selected the empty tile for any unseen state.

\subsubsection{MCMCTS} MCMCTS is an extension of Markov chains introduced by Summerville et al. with the intention of increasing the controllability of generation via MCTS \cite{summerville2015mcmcts}.
It learns the probability distributions for entire columns instead of singular tiles, and then generate columns via an MCTS-driven controller guided by a heuristic that aims to balance playability and the contents of the level.
The chosen heuristic for our implementation measured the difference in the number of pits, enemies, and rewards from the target level, which is loosely based off the described heuristics in the original MCMCTS paper \cite{summerville2015mcmcts}.
We chose MCMCTS as a baseline since it represents the only other prominent use of MCTS within level generation.
Notably, MCMCTS was unable to generate unique levels for the GVGAI Zelda domain, as the levels proved too small to collect a sufficient variety of column-based probabilities to avoid total plagiarism.
Similarly, MCMCTS was unable to generate levels based on $1$-$1$ and $1$-$2$ individually, and as such we present results after training MCMCTS on both levels.

\subsubsection{WFC \& Sturgeon} WFC is a constraint-based PCGML approach which learns constraints for tile placement from example levels, then applies those constraints to greedily generate new levels.
Constraints are learned using a sliding window as in Markov chains, and generation works by repeatedly choosing a random option out of the possibilities for the most constrained tile, then ``collapsing'' any tiles constrained to a single possibility by that choice.
We chose to compare against WFC due to its ability to generate content from a single example.

However, for Super Mario Bros., we found the base implementation of WFC was too slow to generate entire levels at a reasonable rate.
As such, we used Sturgeon, a similar constraint-based PCGML system in WFC's place for the Super Mario Bros. domain \cite{cooper2022sturgeon}.
Sturgeon works similarly to WFC, using a constraint-based pattern matching window while also matching the tile distribution of the source level.
Notably, we chose to avoid the additional global constraint features of Sturgeon such as playability due to the additional human-authoring they represent, and to keep the baseline as close to WFC as possible.
For Super Mario Bros., we utilized a ``ring'' shaped window, which considers each of the $8$ adjacent tiles in learning constraints.
For GVGAI Zelda, we utilized a $2 \times 2$ window due to the relatively smaller size of the levels.
% Notably, while we used the Sturgeon implementation of WFC for our Mario level generation, we did not use any of the global constraint features due to the inherent additional human-authoring they represent \cite{cooper2022sturgeon}.

\subsubsection{Autoencoder} Autoencoders are a type of neural network which are designed to first ``encode'' input data into a compressed representation, and then ``decode'' this data with the goal of reconstructing the input as closely as possible.
In the context of level generation, learning this compressed representation allows us to feed noise into the decoder segment of the autoencoder to generate novel levels.
We chose autoencoders as a baseline due to their success in prior work within Super Mario Bros., along with their ability to train with relatively little data compared to other deep learning models \cite{jain2016autoencoders}.
Our baseline implementation is based off of the autoencoder architecture from Jain et al.'s paper, with minor, domain-dependent changes to the dimensionality.
% architecture details (wonder if it would be easier to include a figure to point to)
% This architecture passes input through a fully-connected encoder layer, followed by a fully-connected latent layer of much lower dimensionality.
% The output of this latent layer is then passed through a fully-connected decoder layer of the same dimensionality as the encoder layer, and finally through an output layer.
This architecture is made up of fully-connected layers only, with one layer for the encoder and decoder.
All layers use ReLU activation, except the output layer of softmax activation to determine the output tokens.
During generation, we input the existing levels to the model with a noise variance of $0.01$ added to each variable.
This scheme, as described in prior work, allows for the generation of levels consistent with the source without direct plagiarism \cite{jain2016autoencoders}.
Further, we used probabilistic decoding rather than a greedy decoding to maximize the variety of output levels.

For Super Mario Bros., we used $16 \times 16$ level chunks as input.
The encoder and decoder layers each had $512$ nodes, and the latent layer consisted of $8$ nodes.
For GVGAI Zelda, we used $1 \times 9$ chunks of level as input, which represent single columns of a level.
The encoder and decoder layers each had $256$ nodes, and the latent layer consisted of $4$ nodes. 
These differences in dimensionality and input size are due to the differences in size between these two domains, as GVGAI Zelda has fewer training examples and potential input states than Super Mario Bros.

Notably, this approach was unable to produce non-empty levels using training data from only a single example in both domains.
We hypothesize that this is due to the model getting trapped in a local minima due to the lack of training examples.
As such, we present results for autoencoders trained on both levels for each domain.

\subsubsection{TOAD-GAN} TOAD-GAN is a level generation model for token-based game levels based on Generative Adversarial Networks (GANs) \cite{awiszus2020toad}.
TOAD-GAN can train on a single input example by downsampling the input level to multiple different scales during training.
We chose to compare against TOAD-GAN due to its ability to generate high-quality levels from only a single example.
In generation, we used pre-trained, open source models provided by the TOAD-GAN authors.

% probably need a reason why we didn't do this for zelda beyond "no time to reimplement." would it struggle to work due to the utter lack of training data? (levels are just 13 * 9 tokens big as opposed to the 16 * 200~ that make up mario levels
%<Some reasoning here about why we didn't implement it for GVGAI Zelda>
Notably, we do not provide results for TOAD-GAN on the Zelda GVGAI domain, as reimplementation and retraining of the approach would require significant design decisions.
While we would ideally be able to compare with this baseline across both domains, we cannot make these design decisions in an unbiased way. 
%While we would ideally be able to compare with this baseline across both domains, the focus of this paper is not on the generalizability of TOAD-GAN, but our own approach.
%Thus we consider the comparison on only the Mario domain sufficient for this initial work.

% overview each metric and why
\subsection{Metrics}
%The metrics used for comparison were playability, plagiarism of the original source levels, and self-similarity between the population of generated levels for each approach.
% motivating why we chose these three metrics, and walking through each one in detail for each domain
To evaluate the ability of each approach to generate a variety of valid, unique levels, we chose playability, plagiarism, and self-similarity.

\begin{table}[]
\begin{tabular}{lccc}
\hline
                      & \textbf{Playable} & \textbf{Plagiarism} & \textbf{Self Sim} \\ \hline
\textbf{TRP-Fixed}    & 95\%                 & 91.21  $\pm$ 0.7           & 94.4 $\pm$ 0.4            \\
\textbf{TRP-Variety}  & 85\%                 & 87.79  $\pm$ 10.5           & 83.3 $\pm$ 8.8            \\
\textbf{Sturgeon}          & 3\%                  & 88.18   $\pm$ 2.6          & 90.1 $\pm$ 3.9            \\
\textbf{MC} & 47\%                 & 81.11   $\pm$ 14.7         & 79.2 $\pm$ 13.1            \\
\textbf{TOAD-GAN}     & 94\%                 & 90.03 $\pm$ 1.0            & 91.3 $\pm$ 1.1            \\ \hline
\end{tabular}
\caption{Results for Super Mario Bros. on Level 1-1.}
\end{table}

\begin{table}[]
\begin{tabular}{lccc}
\hline
                      & \textbf{Playable} & \textbf{Plagiarism} & \textbf{Self Sim} \\ \hline
\textbf{TRP-Fixed}    & 86\%                 & 79.64  $\pm$ 1.5           & 84.9 $\pm$ 0.9           \\
\textbf{TRP-Variety}  & 90\%                 & 78.42  $\pm$ 8.3           & 75.9 $\pm$ 7.1            \\
\textbf{Sturgeon}          & 51\%                 & 68.48 $\pm$ 2.1             & 69.9 $\pm$ 1.0           \\
\textbf{MC} & 14\%                 & 67.63   $\pm$ 3.0          & 66.6 $\pm$ 2.1           \\
\textbf{TOAD-GAN}     & 55\%                 & 82.20    $\pm$ 1.5         & 83.8 $\pm$ 1.3           \\ \hline
\end{tabular}
\caption{Results for Super Mario Bros. on Level 1-2.}
\end{table}

\begin{table}[]
\begin{tabular}{lccc}
\hline
                      & \textbf{Playable} & \textbf{Plagiarism}  & \textbf{Self Sim}   \\ \hline
\textbf{TRP-Fixed}    & 88\% & 74.51 $\pm$ 2.0 & 78.9 $\pm$ 0.8 \\
\textbf{TRP-Variety}  & 88\% & 72.90 $\pm$ 7.4 & 71.4 $\pm$ 7.3 \\
\textbf{MCMCTS}          & 100\%                & 81.23  $\pm$ 1.0           & 85.2 $\pm$ 1.0               \\
\textbf{Autoencoder} & 100\%                & 87.80    $\pm$ 0.2          & 95.6 $\pm$ 0.3             \\ \hline
\end{tabular}
\caption{Results for Super Mario Bros. on both levels.}
\end{table}

We chose playability as a metric as it determines the validity of a generated level.
By validity, we refer to a level being consistent with the global constraints of the game as well as having a solution.
%We intuit that such levels are useful to designers, as they allow for generated levels to be functional without additional human-authoring to fix mistakes or unplayable segments.
In order to measure playability for Super Mario Bros., we used an existing text-based A* playability tester to determine if the ending of the level was reachable from a starting position in a level \cite{summerville2016vglc}.
This starting position was the first available position that was not blocked by solid objects in the third column of the level.
For GVGAI Zelda, we first ensured that the required game elements were present (exactly one player, one or more doors, and one or more keys), and then performed a flood-fill from the player's position to determine if both a key and goal were reachable.

We chose plagiarism as a metric to measure the uniqueness of the levels generated by each approach with respect to the original source levels.
While a certain amount of plagiarism may reflect a generator learning the structure of a level, we intuit that a generator which overtly copies source levels may not be useful to designers.
Further, this metric serves to measure the effect of overfitting on the training data for the machine learning-based baselines.
We used edit-distance to measure the amount of plagiarism between the generated output and source levels, which here is the percentage of tile positions which directly match between the output and source level.
This metric is the same across both domains.

We chose self-similarity as a metric in order to measure the variety of the levels generated by each approach. 
As with plagiarism, we measure self-similarity as an average of the edit-distance of each combination of levels within a generated population.
Self-similarity can be viewed as a measure of the consistency of the output levels from a given generator.
As such, high self-similarity is not necessarily a negative trait, but rather a value can be used to contextualize the performance of a generator.

% clearly stating the goals here right before the results so they can look at the table and know why high number good for playable, but the other two are a little weird
Ideally, a generator would be able to generate a population of levels that were all playable, which minimally plagiarized the source levels, and with a low amount of self-similarity.
However, while it is always strictly better for all levels to be playable, the other two metrics do not indicate quality on their own.
% trying my best to make this point, i think this example is the way to do it? if there's a better way (i was thinking about maybe saying something about how high plagarism might not be bad bc we might be capturing the designer intention but i think i'll just add that in the results section!)
For example, random noise would score low on plagiarism and self-similarity, but would likely be useless to a human designer. 
As such, it is important to consider all three metrics in concert.

\section{Results}
\begin{table}[]
\begin{tabular}{lccc}
\hline
                      & \textbf{Playable} & \textbf{Plagiarism} & \textbf{Self Sim} \\ \hline
\textbf{TRP-Fixed}    & 100\%                & 90.87  $\pm$ 3.4            & 90.81  $\pm$  1.6      \\
\textbf{TRP-Variety}  & 84\%              & 87.11 $\pm$ 7.1            & 84.46 $\pm$ 3.3          \\
\textbf{WFC}          & 0\%                  & 45.83     $\pm$ 5.9         & 50.99  $\pm$ 2.6        \\
\textbf{MC} & 4\%                  & 56.25  $\pm$ 5.7           & 63.61 $\pm$ 3.4          \\ \hline
\end{tabular}
\caption{Results for Zelda GVGAI on Level 1.}
\end{table}

\begin{table}[]
\begin{tabular}{lccc}
\hline
                      & \textbf{Playable} & \textbf{Plagiarism} & \textbf{Self Sim} \\ \hline
\textbf{TRP-Fixed}    & 96\%                 & 88.26  $\pm$ 1.6           & 95.28 $\pm$ 1.1          \\
\textbf{TRP-Variety}  & 72\%                 & 90.44    $\pm$ 4.2        & 89.29 $\pm$ 2.1          \\
\textbf{WFC}          & 0\%                  & 44.99  $\pm$ 5.2            & 48.08 $\pm$ 1.4         \\
\textbf{MC} & 6\%                  & 50.61  $\pm$ 5.5            & 55.08 $\pm$ 1.8         \\ \hline
\end{tabular}
\caption{Results for Zelda GVGAI on Level 2.}
\end{table}

\begin{table}[]
\begin{tabular}{lccc}
\hline
                     & \textbf{Playable} & \textbf{Plagiarism} & \textbf{Self Sim} \\ \hline
\textbf{TRP-Fixed}   & 100\%                & 80.26   $\pm$ 1.9          & 80.30  $\pm$ 1.4          \\
\textbf{TRP-Variety} & 74\%                 & 78.38  $\pm$ 4.6            & 76.58 $\pm$ 2.4        \\
\textbf{Autoencoder} & 18\%                 & 64.32    $\pm$ 1.9          & 83.58 $\pm$ 1.9         \\ \hline
\end{tabular}
\caption{Results for Zelda GVGAI on both levels.}
\end{table}

In this section, we compare the results of TRP and the baseline approaches across both game domains.
% how many levels did we use for comparison, and why?
For Super Mario Bros., we generated 100 levels per approach for comparison, and 50 levels for GVGAI Zelda.
We empirically determined these level counts by observing the highest count that could be reached before our baselines began repeating levels verbatim.
Tables 1, 2 and 3 contain the results of our evaluation for Super Mario Bros., and Tables 4, 5 and 6 contain the results for GVGAI Zelda.
In addition, Figures 2 and 3 depict original and generated levels from both domains.

% what were the takeaways?
From our results across both domains, we identified three major takeaways. \textbf{(i)} On average, TRP generated more playable levels without a hard coded playability constraint, \textbf{(ii)} TRP generated levels with lower source plagiarism and self-similarity than baselines with comparable playability, and \textbf{(iii)} varying TRP's parameters lessens self-similarity and plagiarism at the expense of playability.

% playability
Across both domains, the vast majority of generated levels by TRP were playable.
Notably, playability is not guaranteed by default with TRP due to the binary space partition potentially blocking the paths of the input search trees.
% in mario we beat most but not all
In the Super Mario Bros. domain, TRP outperformed Sturgeon, Markov Chains (MC), and TOAD-GAN overall in playability, consistently generating playable levels regardless of training data.
We hypothesize the changes in playability between $1$-$1$ and $1$-$2$ are due to the more constrained geometry of $1$-$2$, which helps approaches like Sturgeon and hinders approaches like Markov Chains.
% addressing the outlier
Notably, both MCMCTS and the autoencoder approach scored 100\% playability when trained on both levels for Super Mario Bros.
However, similarly to the levels depicted in Figure 2, the autoencoder approach generated largely empty levels, while MCMCTS simply permuted columns from 1-2 until the desired length of level was reached, leading to higher values of plagiarism and self-similarity.
% we crushed zelda
In the GVGAI Zelda domain, TRP vastly outperformed the baseline approaches, which struggled to generate playable levels. 
This is likely due to the hyper-local nature of WFC and MC.
We hypothesize TRP's performance is due to its use of the knowledge kit of goals, which help in ensuring the correct number of required game objects are placed in a generated level.
These results are promising overall, as they show TRP was able to generalize to generate playable levels in both domains.

% TRP generated levels with lower source plagiarism and self-similarity than baselines with comparable playability
Across both domains, TRP was able to achieve lower plagiarism and self-similarity scores than approaches with similar playability.
In particular, TRP-Variety performed well on these metrics, as the variance in parameters allows for a larger variety of generatable levels.
% mario
In the Super Mario Bros. domain, TRP scored substantially lower in plagiarism and self-similarity than TOAD-GAN, MCMCTS, and the autoencoder, which had similar playability to TRP.
These results indicate TRP was able to generate roughly the same amount of playable levels as these approaches while plagiarizing from both the source and itself less.
% zelda (mentioned briefly bc its less relevant here)
While Sturgeon and Markov Chains outperformed TRP according to these metrics in the Super Mario Bros. domain, and overall within the GVGAI Zelda domain, we intuit that this is due to noise based on the unplayable nature of the majority of the levels generated by these approaches. 

% variety in parameters makes a difference
Across both domains, it is clear that the variance of TRP's parameters results in lower plagiarism and self-similarity in comparison to fixed parameters, regardless of the amount of training data.
However, this may result in a drop-off in the playability of generated levels, as seen in the Zelda GVGAI results.
We hypothesize that this is due to the increased variety in the binary space partition step, which has the highest chance of rendering a level unplayable.
More investigation is required to find the best range of values to optimize for playability and against plagiarism for a given domain.

\section{Discussion}

% talk about all the different ways this could go!

% limitations: things that aren't performing perfectly (playability not 100%, mcmcts might beat us out in some cases, plaigirism)

\subsection{Limitations}
% forward model is a pretty big ask
A major limitation of TRP is the requirement of a forward model for generation.
While this is a non-trivial ask for game developers, we hypothesize this issue could be solved if the forward model was already built into a game engine via a plugin.
% playability not 100%
Another limitation of TRP is that playability of generated levels is not strictly guaranteed.
This could potentially be addressed by the addition of a post-processing step in which the existing MCTS agent plays through the final level to ensure playability.
% fixing the parameters results in higher plagiarism
While TRP was able to perform well with both fixed and varied parameters, fixing the parameters resulted in more plagiarism and less variety in generation.
% varying the parameters allieviates this, but can effect playability, and we don't know to what extent
We found varying the parameters alleviates this within our evaluation domains, but this can negatively affect playability of generated levels.
Determining this balance automatically without requiring additional designer input could be a possible solution, which we leave to future work.

\subsection{Future Work}
% we should do a human subject study
An immediate next step would be to utilize TRP in a human-subject study in order to allow designers to develop games alongside this approach.
% what about some auxiliary tasks?
However, there exist several potential applications of TRP to tasks outside of early game development level generation.
An example is level blending, which could be achieved by mixing search trees from multiple source levels during the reconstruction step.
Another is the generation of auxiliary environments for reinforcement learning in order to help with generalizability of agents.
For example, this could entail using TRP to generate levels based on a single source example in order to prepare an agent to generalize across future unseen levels.
% conclusion

\section{Conclusions}
In this paper, we introduced Tree-based Reconstructive Partitioning (TRP), a novel PCGML approach that can generate levels based on only a single example.
We found TRP outperformed other low data PCG approaches across two game domains with respect to playability, plagiarism, and self-similarity.
We consider TRP to be a promising new approach, and hope it can support developers within the early stages of game development without requiring intensive hand-authoring.

\section*{Acknowledgements}

This work was funded by the Canada CIFAR AI Chairs Program. We acknowledge the support of the Natural Sciences and Engineering Research Council of Canada (NSERC).

\bibliography{main}

\end{document}